\documentclass[letterpaper, 10 pt, conference]{ieeeconf}  

\IEEEoverridecommandlockouts                              

\usepackage{graphics} 
\usepackage{epsfig} 
\usepackage{mathptmx} 
\usepackage{times} 
\usepackage{amsmath} 
\usepackage{amssymb}  

\usepackage[pagebackref=true,breaklinks=true,letterpaper=true,colorlinks,bookmarks=false]{hyperref}

\usepackage{graphicx}
\usepackage{url}
\usepackage{booktabs,tabulary,overpic}
\usepackage{multirow}
\usepackage{stfloats}
\usepackage{tabularx}
\usepackage{algorithm}
\usepackage{algpseudocode}
\usepackage{subcaption}
\usepackage{bm}
\usepackage{mathtools}
\usepackage{capt-of}
\usepackage{varwidth}
\usepackage[usenames, dvipsnames]{color}

\definecolor{whitesmoke}{rgb}{0.96, 0.96, 0.96}
\definecolor{antiquewhite}{rgb}{0.98, 0.92, 0.84}
\definecolor{citecolor}{RGB}{34,139,34}

\usepackage{xspace}
\makeatletter
\DeclareRobustCommand\onedot{\futurelet\@let@token\@onedot}
\def\@onedot{\ifx\@let@token.\else.\null\fi\xspace}

\def\eg{\emph{e.g}\onedot} 
\def\ie{\emph{i.e}\onedot} 
 
 \def\vs{\emph{vs}\onedot}

\makeatother

\DeclareMathOperator*{\argmax}{arg\,max}

\title{\LARGE \bf
Deep Mixture Density Network for Probabilistic Object Detection}

\author{Yihui He$^{*}$ and Jianren Wang$^{*}$
\thanks{*indicates equal contribution. Yihui He and Jianren Wang are Robotics Institute, Carnegie Mellon University, 5000 Forbes Ave, PA 15213, USA. {\tt\small \{he2@alumni,jianrenw@andrew\}.cmu.edu}
}}

\begin{document}
\newcommand{\softnms}{soft-NMS~\cite{softnms}}
\newcommand{\ournms}{var voting}
\newcommand{\Ournms}{Variance Voting}
\newcommand{\ourloss}{KL Loss}
\newcommand{\datasetvar}{ambiguity}
\newcommand{\datasetvars}{ambiguities}
\newcommand{\Datasetvars}{Ambiguities}
\newcommand{\modelvar}{uncertainty}
\newcommand{\modelvars}{uncertainties}
\newcommand{\probd}{probabilistic object detection}

\maketitle
\thispagestyle{empty}
\pagestyle{empty}

\begin{figure*}
    \begin{center}
    \includegraphics[width=\linewidth]{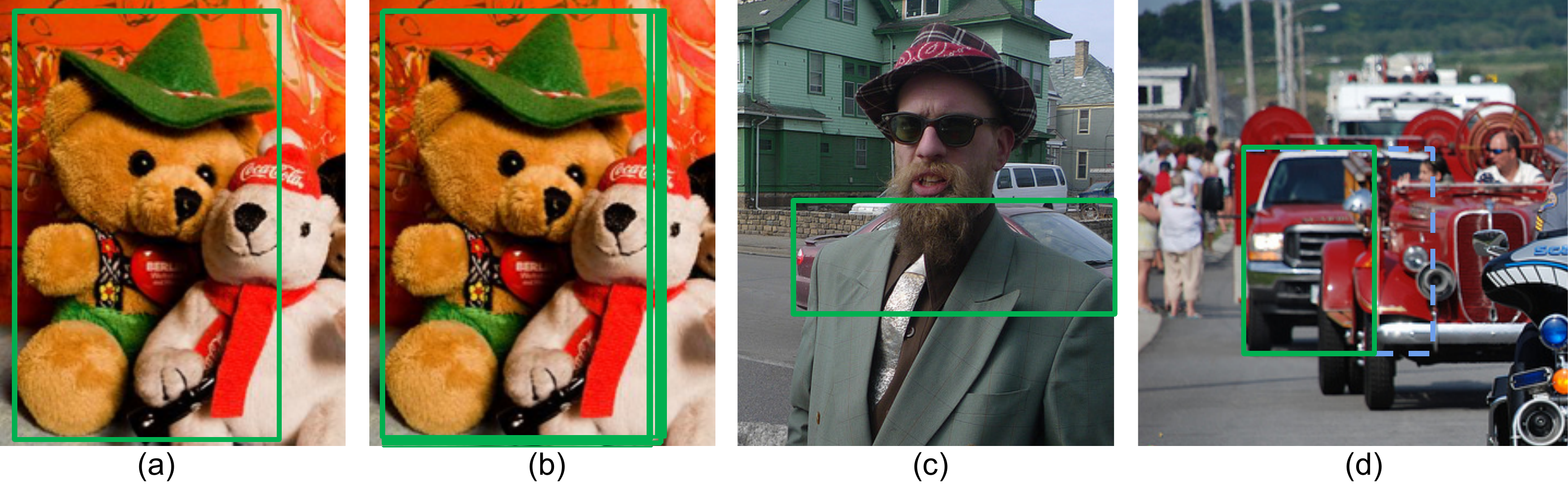}\\
    \caption{We observe that an occluded bounding box usually exhibits multiple modes in most detection datasets, no matter whether the ground truth annotation is visible box or full box: (a) visible bounding box annotation (b) full object bounding box labeled by different annotators (c) visible bounding box annotated accurately (d) visible bounding box annotated inaccurately (\textit{better viewed in color})}
\label{fig:dataset}
\end{center}
\end{figure*}

\begin{abstract}
Mistakes/uncertainties in object detection could lead to catastrophes when deploying robots in the real world. In this paper, we measure the uncertainties of object localization to minimize this kind of risk. Uncertainties emerge upon challenging cases like occlusion. The bounding box borders of an occluded object can have multiple plausible configurations. We propose a deep multivariate mixture of Gaussians model for probabilistic object detection. The covariances  help to learn the relationship between the borders, and the mixture components potentially learn different configurations of an occluded part. Quantitatively, our model improves the AP of the baselines by \textbf{3.9\%} and \textbf{1.4\%} on CrowdHuman and MS-COCO respectively with almost no computational or memory overhead. Qualitatively, our model enjoys explainability since the resulting covariance matrices and the mixture components help measure uncertainties. 
\end{abstract}

\section{Introduction}

Object detection provides crucial perception information for real-world applications like robotics grasping~\cite{morethanafeeling} and self-driving cars~\cite{huval2015empirical}. Mistakes/uncertainties in object detection could potentially risk the success of the robot's operation and even human lives~\cite{amodei2016concrete,varshney2017safety,hall2020probabilistic}. In this paper, we minimize this kind of risk by measuring uncertainties of bounding box localization. The uncertainties are quantified by a simple Gaussian mixture model.


Currently, there are two styles of bounding box annotation among the large-scale object detection datasets: (1) visible box that only contains visible parts (\eg, MS-COCO~\cite{coco} and PASCAL VOC~\cite{voc}) (2) full box that contains both visible and occluded parts (\eg, CrowdHuman~\cite{crowdhuman} and VehicleOcclusion~\cite{vehicleocclusion}). For full box annotation, regressing a single set of bounding box coordinates works well for fully visible objects, since it is a unimodal problem. However, when an object is occluded, we observe that its occluded parts can have several plausible configurations (\eg, Figure~\ref{fig:dataset}~(b)), which is a multimodal problem. Even for visible box annotation, an object sometimes still exhibits multiple modes due to inaccurate labeling (\eg, Figure~\ref{fig:dataset}~(c) \vs (d)). We argue that an object detector robust to occlusion should learn a multimodal distribution with the capability of proposing more than one plausible hypothesis for the configuration of an occluded part. 

Besides, we also observe that the bounding box coordinates have correlations by nature. Take Figure~\ref{fig:dataset}~(c) as an example, by knowing the position of the car's roof, we can easily infer the location of the left border even without looking at it. Therefore, an object detector robust to occlusion also needs to be capable of inferring the correlations between the occluded bounding box borders and the visible ones.    

Motivated by these two observations, we propose a deep multivariate mixture of Gaussians model for object detection under occlusion. Concretely, instead of regressing a single set of bounding box coordinates, our model regresses several sets of coordinates, which are the means of the Gaussians. Moreover, we learn a covariance matrix for the coordinates of each Gaussian mixture component. These components are summed together as the prediction for the distribution of plausible bounding box configurations. At inference time, we choose the expectation of our model's distribution as the final predicted bounding box. 

To demonstrate the generalizability of our proposed model, we conduct experiments on four datasets: CrowdHuman, MS-COCO, VehicleOcclusion, and PASCAL VOC. Quantitatively, our model improves the AP (Average Precision of the baselines by \textbf{3.9\%} and \textbf{1.4\%} on CrowdHuman and MS-COCO respectively (Table~\ref{tab:crowdhuman} and Table~\ref{tab:coco}). Qualitatively, our model enjoys explainability since the resulting bounding boxes can be interpreted using the covariance matrices and the Gaussian mixture components (Figure~\ref{fig:sigma} and Figure~\ref{fig:mix}). More importantly, our model is almost computation and memory free, since predicting the mixture components only requires a fully-connected layer, and we can discard the covariance matrices at inference time (Table~\ref{tab:time}).

\section{Related Work}
\paragraph{Object Detection} Deep convolutional neural networks were first introduced to object detection in R-CNN~\cite{rcnn} and Fast R-CNN~\cite{fast}. Currently, there are mainly two types of object detectors: one-stage object detectors and two-stage object detectors. One-stage detectors like YOLO~\cite{yolo}, SSD~\cite{ssd} and RetinaNet~\cite{fl} are fast in general. Two-stage detectors~\cite{faster,mask,dcnv2,sniper} are accurate however sacrificing speed. Although, we conduct most experiments based on the Faster R-CNN heads of Faster R-CNN and Mask R-CNN, we also find that our method consistently improves single stage methods like YOLOv3~\cite{yolov3} (Table~\ref{tab:vocvgg}). 

\paragraph{Object Detection Under Occlusion} Occlusion-aware R-CNN~\cite{orcnn} proposes to divide pedestrian detection into five parts and predict the visibility scores respectively, which are integrated with the prior structure information of the human body into the network to handle occlusion. \cite{zhang2018occluded} proposes an attention network with self or external guidance. These methods are specifically designed for pedestrian detection task. By contrast, our method is designed for general object detection. 

Deep Voting~\cite{deepvoting} proposes to utilize spatial information between visual cues and semantic parts and also learn visual cues from the context outside an object. However, detecting semantic parts needs manual labels, which our approach does not require. Besides, our approach does not introduce additional computation during the inference (Table~\ref{tab:time}). Amodal instance segmentation~\cite{amodal} considers the task of predicting the region encompassing both visible and occluded parts of an object. The authors propose to add synthetic occlusion to visible objects and retain their original masks, then employ a CNN to learn on the generated composite images, which resembles the VehicleOcclusion in our experiments. Based on our previous work \cite{klloss} propose bounding box regression with uncertainty, which is a degradation case of our model (Gaussian). 

\paragraph{Datasets for Detection under Occlusion} Currently, there are three categories of annotation for an occluded object: (1) visible bounding box that contains the visible parts (2) full box that contains both visible and occluded parts of an object annotated by human (3) full box by synthesizing occluders on a visible object. MS-COCO, PASCAL VOC, ImageNet~\cite{imagenet} and Cityscapes~\cite{cityscapes} fall into the first category. CrowdHuman and Semantic Amodal Segmentation dataset~\cite{semanticamodal} require the annotators to label the invisible parts. VehicleOcclusion instead synthesizes the occluders for visible objects. We conduct experiments on MS-COCO, PASCAL VOC, CrowdHuman, and VehicleOcclusion, covering all these categories.
\section{Approach}

\begin{figure}[t]
\begin{center}
\includegraphics[width=.9\linewidth]{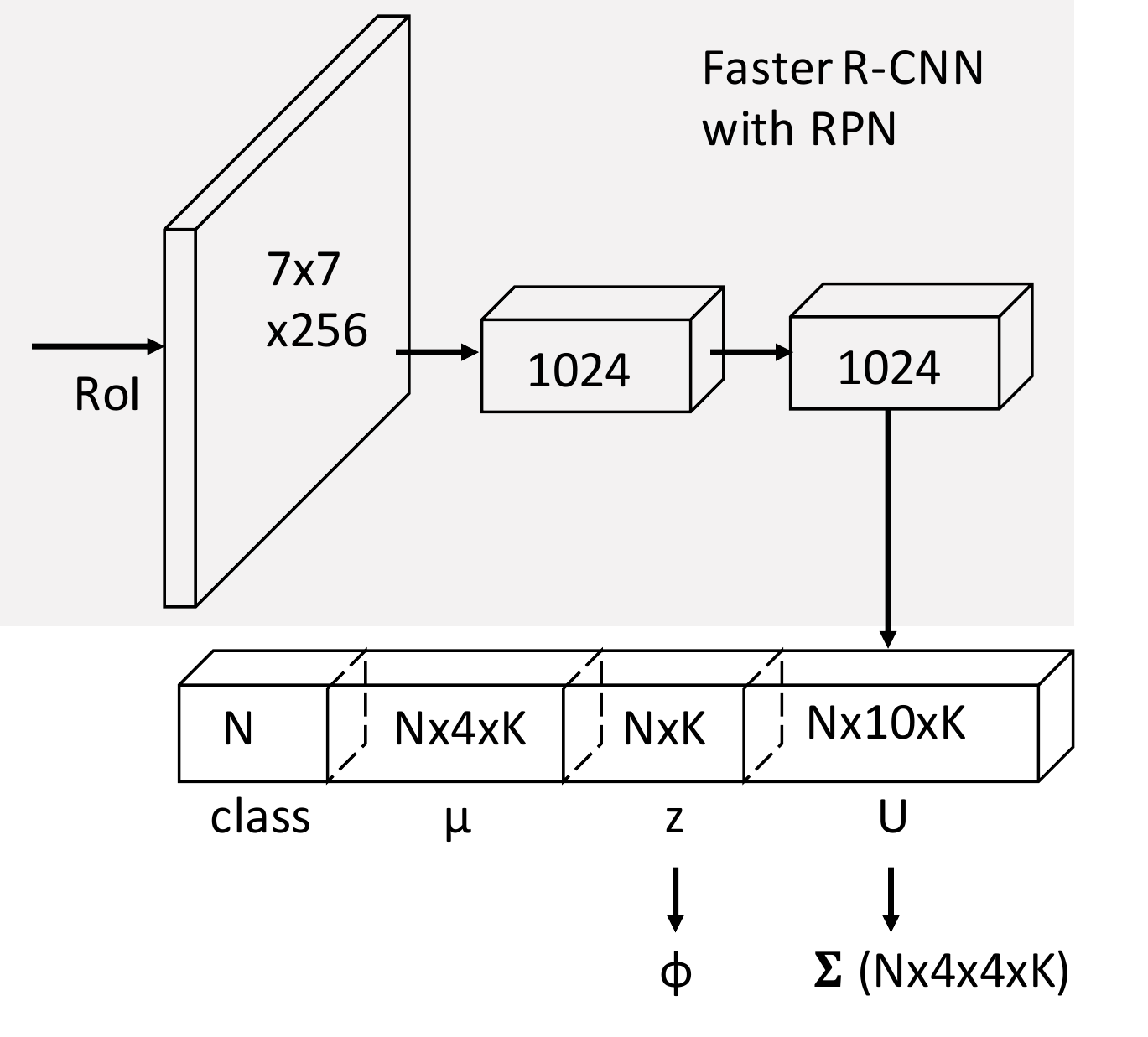}
\caption{Faster R-CNN head architecture for our approach: We extended the existing Faster R-CNN head to predict the parameters of multivariate mixture of Gaussian $\bm{\mu}$, $\bm{\phi}$ and $\bm{\Sigma}$}
\label{fig:arch}
\end{center}
\end{figure}

\subsection{Formulation}\label{sec:inference}
\newcommand{\roi}{\bm{I}}
We observe that when an object is partially occluded, the occluded bounding box border can usually be inferred to some extent by other visible parts of the object (\eg, it is easy to infer the left border of the car given the car roof position in Figure~\ref{fig:dataset} (c)). Besides, the occluded bounding box exhibits multiple modes. For example, the left arm of the teddy bear could have several possible configurations in Figure~\ref{fig:dataset}~(b). Motivated by these two observations, we propose to estimate the bounding box coordinates as a probability distribution during bounding box regression instead of a set of deterministic coordinates. Specifically, we propose to estimate a multivariate mixture of Gaussians distribution with a deep network~\cite{bishop1994mixture}. Multivariate Gaussian helps the case where bounding box borders have correlations, and a mixture of Gaussians helps the case where an occluded bounding box border exhibits multiple modes. Formally, we predict the distribution $p_\theta(\bm{x}|\roi)$ given the feature maps $\roi$ of a region of interest (RoI). The distribution is parameterized by $\theta$, which is a neural network (\eg, Faster R-CNN head, Figure~\ref{fig:arch}). The distribution has $K$ components $\mathcal{N}(\bm{\mu_i}, \bm{\Sigma_i})$. Each component $i$ has mean $\bm{\mu}_{i=1...K} = [x_1, y_1, x_2, y_2]^T$, which is the most probable bounding box coordinates relative to the RoI, estimated by the component:

\begin{equation}\label{eq:model}
\begin{aligned}
    p_\theta(\bm{x} | \roi) &= \sum_{i=1}^K \phi_i \mathcal{N}(\bm{\mu_i}, \bm{\Sigma_i})  \\
    \text{   where }  & \sum_{i=1}^K \phi_i = 1  \text{ and } 0\le \phi_i \le 1\\
    \mathcal{N}(\bm{\mu_i}, \bm{\Sigma_i}) &= \dfrac{ \exp\Big(-\dfrac{1}{2}(\bm{x}-\bm{\mu_i})^T\bm{\Sigma_i}^{-1}(\bm{x}-\bm{\mu_i})\Big)}{\sqrt{(2\pi)^4\lvert \bm{\Sigma_i}\rvert}} \\
\end{aligned}
\end{equation}

where each $\mathcal{N}(\bm{\mu_i}, \bm{\Sigma_i})$ is a multivariate Gaussian distribution. $\phi_i$ is a  mixture weight scalar for $\mathcal{N}(\bm{\mu_i}, \bm{\Sigma_i})$. $\lvert \bm{\Sigma_i} \rvert$ is the determinant of $\bm{\Sigma_i}$. $\bm{\Sigma}$ is the covariance matrix, which is a symmetric semi-positive definite matrix in general. To be able to compute the inverse $\bm{\Sigma}^{-1}$, we constrain the covariance matrix to be a symmetric positive definite matrix. In this case, the precision matrix $\bm{\Sigma}^{-1}$ is also a symmetric positive definite matrix. During training, the model estimates the precision matrix $\bm{\Sigma}^{-1}$ instead of the covariance matrix $\bm{\Sigma}$, so that we do not need to compute the inverse every time during training which we also find more stable in our experiments.
To ensure the properties of the precision matrix $\bm{\Sigma}^{-1}$, we parameterize it using the Cholesky decomposition: 
\begin{equation}\label{eq:sigma}
\begin{aligned}
\bm{\Sigma}^{-1} &= \bm{U}^T\bm{U} \\
\end{aligned}
\end{equation}
where $\bm{U}$ is an upper triangular matrix with strictly positive diagonal entries, such that Cholesky decomposition is guaranteed to be unique:
\begin{equation}
\begin{aligned}
\bm{U}&= \begin{bmatrix}
 \exp(u_{11}) & u_{12} & u_{13} & u_{14} \\ 
 &   \exp(u_{22}) & u_{23} & u_{24}\\ 
 &  &   \exp(u_{33}) & u_{34}\\ 
 &  &  &   \exp(u_{44})
\end{bmatrix}\\
\lvert \bm{\Sigma} \rvert 
&= \dfrac{1}{\lvert \bm{\Sigma}^{-1} \rvert} 
= \dfrac{1}{\lvert \bm{U}^T\bm{U} \rvert} 
= \dfrac{1}{\lvert \bm{U}^T \rvert\lvert \bm{U} \rvert} 
= \dfrac{1}{\lvert \bm{U} \rvert^2} \\ &=  \dfrac{1}{\exp(\sum_{i=1}^4u_{ii})^2} \\
\end{aligned}
\end{equation}

We parameterize the mixture weights $\phi_i$ using Softmax, 
so that they range from 0 to 1 and sum to 1:
\begin{equation}\label{eq:phi}
\begin{aligned}
    \phi_i = \dfrac{\exp(z_i)}{\sum_{k=1}^K \exp(z_k)}
\end{aligned}
\end{equation}

$z_i$, $u_{ii}$ and $\mu_i$  are outputs produced by a fully-connected layer on top of the final fully-connected layer fc7 on the Faster R-CNN head. Take Faster R-CNN with RPN as an example, Figure~\ref{fig:arch} shows the architecture of our model. Since we only modify a small part of the architecture, our approach might also be applied to other object detectors than Faster R-CNN, like one-stage object detectors YOLO and RetinaNet.

\paragraph{Learning} Our model parameterizes the distribution over bounding boxes using a neural network which depends on RoI features. During training, we estimate the parameters $\theta$ with maximum likelihood estimation on a given dataset $\{\roi_\ell,\ \bm{\mu}^{*}_\ell | \ell = 1,2,...,N\}$, where $\bm{\mu}^{*}_\ell$ represents the ground truth coordinates for RoI feature maps $\roi_\ell$ and $N$ is the number of observations: 
\begin{equation}
\begin{aligned}
\hat{\theta} = \argmax_\theta \dfrac{1}{N}\sum_\ell^N \ln p_\theta(\bm{\mu}^*_\ell | \roi_\ell)
\end{aligned}
\end{equation}

In practice, $N$ is the number of samples in a mini-batch. We use momentum stochastic gradient descent (SGD) to minimize the localization loss $L_{loc}$ and the classification loss $L_{cls}$:
\begin{equation}
\begin{aligned}
L &= L_{cls} + \lambda L_{loc} \\ 
\text{   where }  & L_{loc} = - \dfrac{1}{N}\sum_\ell^N \ln p_\theta(\bm{\mu}^*_\ell | \roi_\ell) \\
\end{aligned}
\end{equation}

Note that we use different parameters $\theta$ for different classes in practice. For simplicity, the formulation above only considers the regression problem for a single class. 

\paragraph{Inference} 
During testing, we perform inference in two ways: The most obvious one is \textbf{average}. we use the expectation of our mixture module as prediction: 
\begin{equation}\label{eq:expectation}
    \mathbb{E} \left [p_\theta(\bm{x} | \roi)\right ] = \dfrac{1}{K}\sum_{i=1}^K\phi_i\bm{\mu_i}
\end{equation}

Notice that the covariance matrix $\bm{\Sigma}_i$ is not involved in inference. In practice, we discard the neurons that produce the covariance matrix to speed up inference. In our experiments (Table~\ref{tab:time}), our model has almost the same inference latency and memory consumption as the baseline network.

The second way is most \textbf{probable} inference, where we use the prediction from the mixture component with highest confidence:
\begin{equation}\label{eq:probable}
     \bm{\mu_i} \text{ where } i = \argmax_i \phi_i
\end{equation}
This is not robust in practice. We propose to use the inference when the $\phi_i$ from the most confident component ($i$) is highest than $\phi_j+ t$ from the second most confident component ($j$). $t > 0$ is a threshold, (\eg, $0.4$). If this is not satisfied, we use the average inference.

\subsection{Degradation Cases}
\paragraph{Multivariate Gaussian} When the number of mixture components $K=1$, our model degrades into a multivariate Gaussian model. And the localization loss can be rewritten as follow (for simplicity, we only illustrate the loss for a single sample $\ell$):
\begin{equation}\label{eq:multivariate}
\begin{aligned}
L_{loc}^\ell &= \dfrac{(\bm{\mu}^{*}-\bm{\mu})^T\bm{\Sigma}^{-1}(\bm{\mu}^{*}-\bm{\mu})}{2}  + \dfrac{\log\lvert \bm{\Sigma}\rvert}{2} + 2\ln 2\pi \\
&= \dfrac{(\bm{\mu^{*}}-\bm{\mu})^T\bm{U}^T\bm{U}(\bm{\mu^{*}}-\bm{\mu})}{2}  - \sum_{i=1}^4u_{ii} + 2\ln 2\pi  \\ 
\end{aligned}
\end{equation}

where $2\ln 2\pi$ is a constant which can be ignored during training. Multivariate Gaussian model is helpful under occlusion since the borders of a bounding box have correlations with each other inherently. For example, by looking at the location of a car's door, we can guess the location of its roof even if it is occluded. 

\paragraph{Mixture of Gaussians} When the covariance matrix is constrained to be a diagonal matrix, our model becomes a mixture of Gaussians model with independent variables:
\begin{equation}\label{eq:mix}
\begin{aligned}
   L_{loc}^\ell & = -\ln\sum_{i=1} ^K \phi_i\sum_{j=1} ^4\dfrac{\exp\Big(-(\bm{\mu}^{*}_{ij}-\bm{\mu}_{ij})^2/2(\bm{U_i})_{jj}^2\Big)}{\sqrt{2\pi}\bm{(U_i)}_{jj}}
\end{aligned}
\end{equation}

where $\bm{(U_i)}_{jj}$ is the $j$th diagonal element of the matrix $\bm{U_i}$. Multimodality is helpful under occlusion because an occluded object usually has multiple modes. 

\paragraph{Gaussian} When the number of mixture components $K=1$ and the covariance is constrained to be a diagonal matrix, it becomes a simple Gaussian model where different variables are independent:
\begin{equation}\label{eq:single}
\begin{aligned}
   L_{loc}^\ell & =\sum_{j=1} ^4 (\bm{U})_{jj}^2\dfrac{(\bm{\mu}^{*}_j-\bm{\mu}_j)^2}{2} - \ln\bm{(U)}_{jj}+\dfrac{\ln 2\pi}{2}
\end{aligned}
\end{equation}

We argue that this simple model helps detection in most cases. Here $\bm{(U)}_{jj}$ behaves like a balancing term. When the bounding box regression is inaccurate (large $(\bm{\mu}^*_j-\bm{\mu}_j)^2/2$), the variance $1/\bm{(U)}^2_{jj}$ tends to be larger. Therefore smaller gradient will be provided to bounding box regression $(\bm{U})_{jj}^2(\bm{\mu}^*_j-\bm{\mu}_j)^2/2$ in this case, which might help training the network (Table~\ref{tab:crowdhuman} and Table~\ref{tab:coco}). If bounding box regression is perfect, $\bm{U}$ tend to infinity (\ie, the variance should be close 0). However, regression is not that accurate in practice, $\bm{U}$ will be punished for being too large. 

\paragraph{Euclidean Loss} When all the diagonal elements $(\bm{U})_{jj}$ are one ($u_{jj} = 0$), our model degenerates to the standard euclidean loss:
\begin{equation}
\begin{aligned}
   L_{loc}^\ell & =\sum_{j=1} ^4 \dfrac{(\bm{\mu}^*_j-\bm{\mu}_j)^2}{2}+\dfrac{\ln 2\pi}{2}
\end{aligned}
\end{equation}



\section{Experiments}
We initialize the weights of $\bm{\mu}_i$, $z_i$ and $u_{ii}$ layers (Figure~\ref{fig:arch}) using random Gaussian initialization with standard deviations $0.0001$ and biases $0$, $-1$ and $0$ respectively. So that at the start of training, bounding box coordinate $\bm{\mu}_i$ is at an unbiased position, $\bm{U}_i$ is an identity matrix and $\phi_i$ treats each mixture component equally. Our model can be trained end-to-end. Unless specified, we follow settings in Detectron~\cite{detectron} and those original papers.


\begin{table}[t]
\begin{center}
 \caption{Performance of our models on CrowdHuman on ResNet-50 FPN Faster R-CNN}
\begin{tabular}{l|c}
method & mAP \\ \hline
RetinaNet~\cite{fl} &80.8 \\
FPN~\cite{fpn} & 85.0 \\ \hline
FPN+Gaussian  & 86.2 \\ 
FPN+mixture of 8 Gaussian (average)  & 87.7 \\ 
FPN+mixture of 8 Gaussian (probable)  & 87.8 \\ 
FPN+multivariate Gaussian & 88.3 \\
FPN+multivariate mixture of 8 Gaussian (average) & 88.8 \\ 
FPN+multivariate mixture of 8 Gaussian (probable) & \textbf{88.9} \\ 
\end{tabular}
\label{tab:crowdhuman}
\end{center}
\end{table}

\begin{table*}[t]
\begin{center}
\caption{Performance of our models on MS-COCO on ResNeXt-101 FPN Faster R-CNN}
\begin{tabular}{l|c|c|c|c|c|c}
method  & AP & AP$^{50}$ & AP$^{75}$ & AP$^S$ & AP$^M$ & AP$^L$  \\ \hline 
baseline &          40.8&62.5&44.4&23.4&44.4&53.8\\
Gaussian &  41.2&61.2&44.3&23.0&44.7&54.4\\
mixture of 8 Gaussian (average) &  41.4&61.2&44.7&23.6&44.6&55.6\\
mixture of 8 Gaussian (probable) &41.6&61.2&44.9&23.6&44.7&55.5\\
multivariate Gaussian &  41.5&61.5&44.7&23.6&45.0&55.2\\
multivariate mixture of 8 Gaussian (average) &42.0&61.6&45.2&23.7&45.1&56.0\\
multivariate mixture of 8 Gaussian (probable) & 42.3&61.6&45.4&23.8&45.2&56.2\\
    \end{tabular} 
\label{tab:coco}
\end{center}
\end{table*}

\begin{table}[tb]
\caption{Probabilistic object detectors comparison. Legend: moLRP = Localization Recall Precision~\cite{oksuz2018localization}, PDQ = Probability-based Detection Quality measure~\cite{hall2020probabilistic}.}
\begin{center}
\begin{tabular}{@{}l|rrr@{}}
Approach ($\tau$) & mAP & moLRP & PDQ  \\ 
 & (\%) & (\%) & (\%)\\
\hline
MC-Dropout SSD (0.5) \cite{miller2018evaluating} & 15.8 & 84.4 & 12.8 \\	
MC-Dropout SSD (0.05) \cite{miller2018evaluating} & 19.5 & 83.4 & 1.3 \\	
SSD-300 (0.5) \cite{ssd} & 15.0 & 85.7 & 3.9 \\	
SSD-300 (0.05) \cite{ssd} & 19.3 & 84.0 & 0.6 \\ 
YOLOv3 (0.5) \cite{yolov3} & 29.7 & 69.2 & 5.7\\	
YOLOv3 (0.05) \cite{yolov3} & 30.1 & 72.3 & 3.3\\	
FRCNN R (0.5) \cite{faster} & 32.8 & 70.9 & 6.7 \\
FRCNN R (0.05) \cite{faster} & 34.3 & 70.9 & 3.0 \\
FRCNN R+FPN (0.5) \cite{massa2018mrcnn} & 34.6 & 68.8 & 11.8 \\
FRCNN R+FPN (0.05) \cite{massa2018mrcnn} & 37.0 & 69.6 & 4.2\\
FRCNN X+FPN (0.5) \cite{massa2018mrcnn} & 37.4 & 67.3 & 11.9\\
FRCNN X+FPN (0.05) \cite{massa2018mrcnn} & 39.0 & 67.9 & 4.4\\
probFRCNN (0.5)~\cite{hall2020probabilistic} & 35.5 & 67.8  & 29.4 \\ \hline
FRCNN+multivariate Gaussian (probable) & \textbf{41.6} &  \textbf{65.3} & \textbf{28.2}
\end{tabular}
\end{center}
\label{tab:evaluation}
\end{table}

\subsection{Datasets}
To demonstrate the generalizability of our method, we conduct experiments on four datasets:

\paragraph{CrowdHuman}~\cite{crowdhuman} is a large, rich-annotated and highly diverse dataset for better evaluation of detectors in crowd scenarios. Its training and validation sets contain a total of 470k human instances, and around 22.6 persons every image under various kinds of occlusions. The annotations for occluded bounding boxes are full boxes (Figure~\ref{fig:dataset}~(b)) instead of visible boxes (Figure~\ref{fig:dataset}~(a)). The experiments are in Table~\ref{tab:crowdhuman}.

\paragraph{VehicleOcclusion} is a synthetic dataset designed for object detection under occlusion~\cite{vehicleocclusion}. Same as above, the annotations are full boxes. The occlusion annotations are more accurate since the occluders (occluding objects) are randomly placed on the annotated visible object. It contains six types of vehicles and occluded instances at various difficulty levels. Specifically, it consists of four occlusion levels: No occlusion ($0\%$), L1 ($20\% \sim 40\%$), L2 ($40\% \sim 60\%$), L3 ($60\% \sim 80\%$). The percentages are computed by pixels. At level L1, L2 and L3, there are two, three, and four occluders placed on the object, respectively (Table~\ref{tab:deepvoting}). 

\paragraph{MS-COCO}~\cite{coco} is a large-scale object detection dataset containing 80 object categories, $330k$ images ($>200k$ labeled) and 1.5 million object instances. Compared with the two datasets above, MS-COCO has fewer occlusion cases. For example, the IoU (intersection over union) between overlapped human bounding boxes in MS-COCO are less than 0.7~\cite{crowdhuman}. We use train2017 for training and val2017 for testing (Table~\ref{tab:coco}). Different from above, the annotations are visible boxes. 

\paragraph{PASCAL VOC} is a classic dataset for object detection~\cite{pascal-voc-2007}. Similar with MS-COCO, this dataset has less occlusion cases than the first two datasets. We use voc\_2007 and voc\_2012 for training and voc\_2007\_test for testing (Table~\ref{tab:vocvgg}).  The annotations are visible boxes.

\subsection{Ablation Study}

\begin{table}[t]
\begin{center}
 \caption{Self-comparison on PASCAL VOC with YOLO-v3}
\begin{tabular}{l|c}
method & mAP \\ \hline
baseline & 79.3 \\ \hline 
Gaussian  & 80.8 \\ 
mixture of 8 Gaussian (average)  & 80.9 \\ 
mixture of 8 Gaussian (probable)  & 81.1 \\ 
multivariate Gaussian & 81.2 \\
multivariate mixture of 8 Gaussian (average) & 81.3 \\ 
multivariate mixture of 8 Gaussian (probable) & \textbf{81.5} \\ 
\end{tabular}
\label{tab:vocvgg}
\end{center}
\end{table}

\begin{figure}
\begin{center}
    \includegraphics[width=\linewidth]{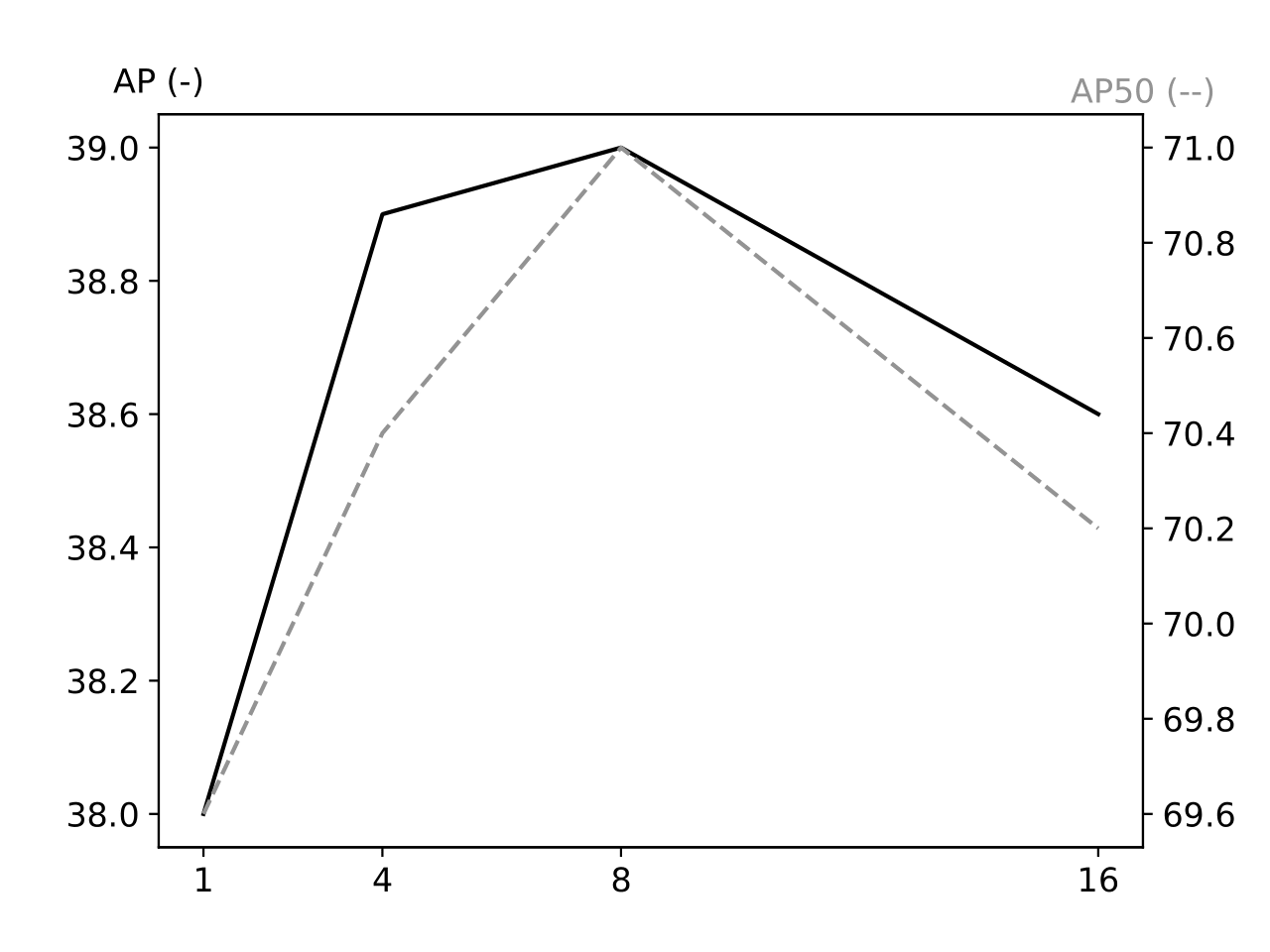}
    \caption{AP and AP50 when varying the number of mixture components}
    \label{fig:nmix}
\end{center}
\end{figure}

\begin{figure}
\begin{center}
    \includegraphics[width=\linewidth]{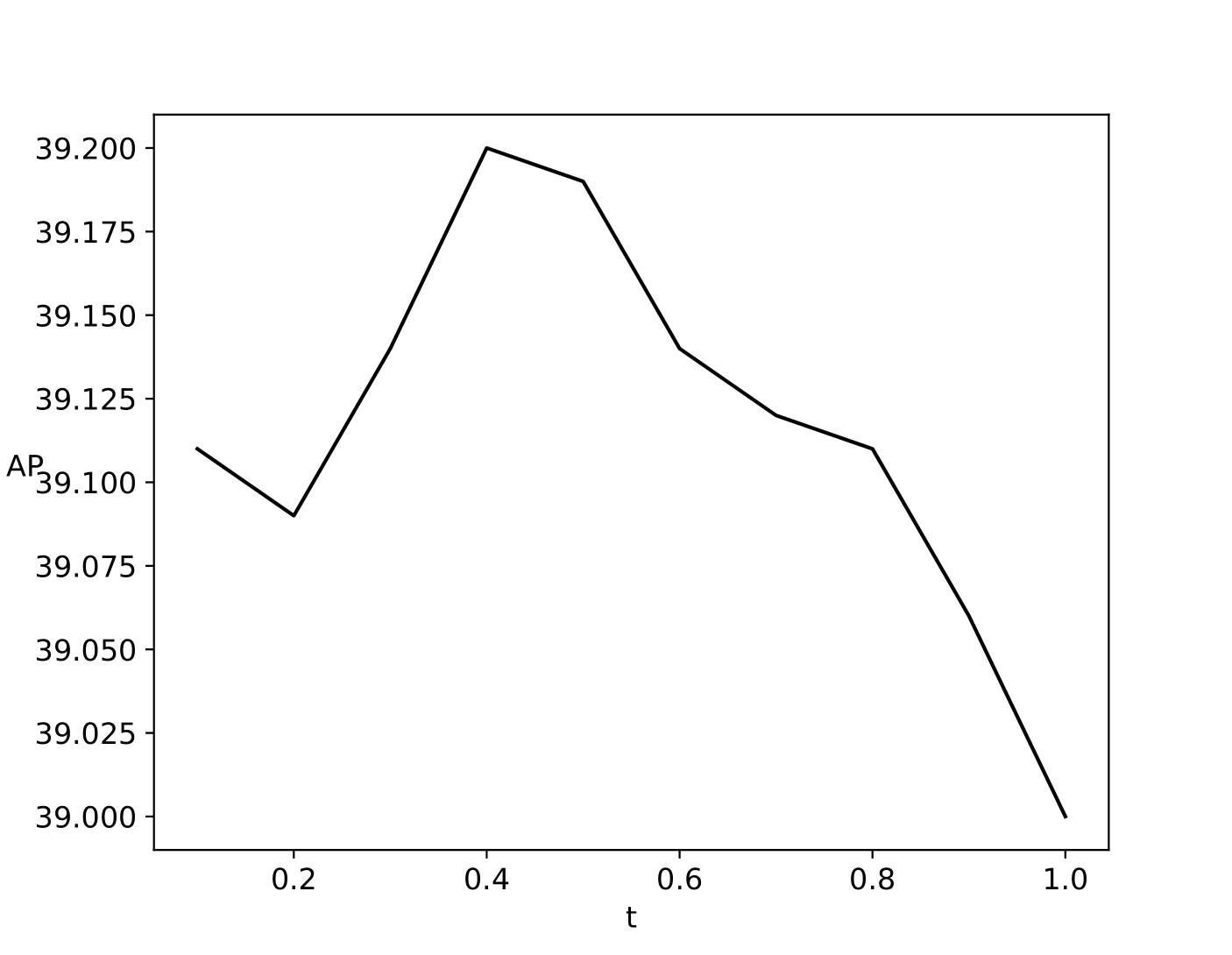}
    \caption{AP by varying the threshold $t$ of probable inference}
    \label{fig:probable}
\end{center}
\end{figure}

\paragraph{Number of Mixture Components} Shown in Figure~\ref{fig:nmix}, we test our mixture of Gaussians model by varying the number of mixture components. The baseline is ResNet-50 FPN Faster R-CNN~\cite{resnet,fpn} on CrowdHuman. As the number of components increases from 1, 4 to 8, we observe consistent performance improvement. The mixture of eight Gaussians model (Equation~\ref{eq:mix}) outperforms Gaussian model (Equation~\ref{eq:single}) by \textbf{1\%} AP. However, the performance goes down when there are more than 16 components. This might be because the objects in the dataset might not have as many as 16 modes when occluded. Besides, the more components we have, the higher the chance of over-fitting. Unless specified, we use eight components for the mixture of Gaussians model.

\paragraph{Average v.s. Probable} Shown in Figure~\ref{fig:probable}, we test ResNet-50 FPN Faster R-CNN mixture of eight Gaussians model with probable inference by varying its threshold $t$. The threshold indeed has a sweet spot. We used $t=0.4$ by default for probable inference for the rest of our paper. Not shown on the figure, when $t=0$, probable inference is very unstable, which only achieves 33.45\%. Also shown in Table~\ref{tab:crowdhuman} and Table~\ref{tab:coco}, probable inference consistently performs better than average inference. 

\paragraph{Mixture of Gaussian \vs Multivariate Gaussian} Shown in Table~\ref{tab:crowdhuman} and \ref{tab:coco}, we compare the degradation cases of our complete model (Equation~\ref{eq:model}): Gaussian (Equation~\ref{eq:single}), mixture of Gaussians (Equation~\ref{eq:mix}) and multivariate Gaussian (Equation~\ref{eq:multivariate}) on CrowdHuman and MS-COCO. For CrowdHuman, we use ResNet-50 FPN Faster R-CNN as the baseline. For MS-COCO, we use ResNeXt-101 (64x4d) FPN Faster R-CNN. 

On CrowdHuman which has a lot of crowded scenes, our model greatly improves the baseline. Gaussian improves the baseline by \textbf{1.2\%} mAP. A mixture of eight Gaussians improves \textbf{2.7\%} mAP, and multivariate Gaussians improves \textbf{3.3\%} mAP. The complete model improves the performance by \textbf{3.9\%} mAP. The improvements indicate all these assumptions are helpful under heavy occlusion. Gaussian helps training the regression network by learning to decrease the gradients for high variance cases. Multivariate Gaussian helps to learn the correlations between an occluded border and the visible borders. Mixture of Gaussians helps to learn a multimodal model for the occluded cases which have multiple modes. 

On MS-COCO, the bounding box annotations are visible boxes instead of full boxes used in CrowdHuman. Gaussian still works here which improves the baseline by \textbf{0.4\%} AP, since there are variances in the dataset caused by inaccurate annotation (\eg, Figure~\ref{fig:dataset}~(d)). Gaussian helps to reduce the gradients for these ambiguous cases. A mixture of eight Gaussians improves \textbf{0.8\%} AP, and multivariate Gaussians improves \textbf{0.7\%} AP. The complete model improves the performance by  \textbf{1.4\%} AP. The improvements are noticeable, however less significant than on CrowdHuman. On the one hand, there are fewer occluded instances in MS-COCO, multimodality and covariances might not be as helpful as in CrowdHuman. On the other hand, predicting full boxes require guessing the invisible parts where multimodality and covariances are more useful.

We further conduct experiments on PASCAL VOC, shown in Table~\ref{tab:vocvgg}. YOLO-v3~\cite{yolov3} is the baseline. Similar to MS-COCO, the bounding box annotations are visible boxes instead of full boxes used in CrowdHuman. We observe that Gaussian improve the mAP (mean Average Precision) by \textbf{1.5\%}. The complete model improves the mAP by \textbf{2.2\%}. Multimodality and multivariate Gaussian do not substantially improve the performance. These observations coincide with the observations on MS-COCO. 
\begin{table}
\begin{center}
\caption{Comparison with a state-of-the-art occlusion-aware detector on  VehicleOcclusion. The metric is mAP. occ.: occlusion}
\begin{tabular}{c|c|c|c|c}
         & no occ. & L1 & L2 & L3  \\ \hline
baseline & 73.6 & 48.3 & 35.0 & 23.0 \\ 
DeepVoting & 72.0 & 53.7 &42.6 & 31.6 \\ 
DeepVoting+ & 74.0 & 58.0 &46.9 & 35.2 \\ \hline 
Ours (average) & \textbf{74.4} & 62.1 & 50.9 & 38.4 \\
Ours (probable) & \textbf{74.4} & \textbf{62.2} & \textbf{51.0} & \textbf{38.6} \\
\end{tabular}
\label{tab:deepvoting}
\end{center}
\end{table}

\begin{table}[t]
\begin{center}
\caption{Model size and FPS comparison with ResNet-50 FPN Mask R-CNN on a single GPU}
\begin{tabular}{l|c|c}
method & \# params & FPS  \\ \hline
baseline & 91M  & 11.1\\  \hline 
Gaussian & 91M  & 11.1\\
multivariate Gaussian & 91M  & 11.1\\
mixture of 8 Gaussian & 93M  & 10.2\\ 
multivariate mixture of 8 & 93M  & 10.2\\ 
\end{tabular}
\label{tab:time}
\end{center}
\end{table}

\begin{figure*}[t]
\begin{center}
    \includegraphics[width=\linewidth]{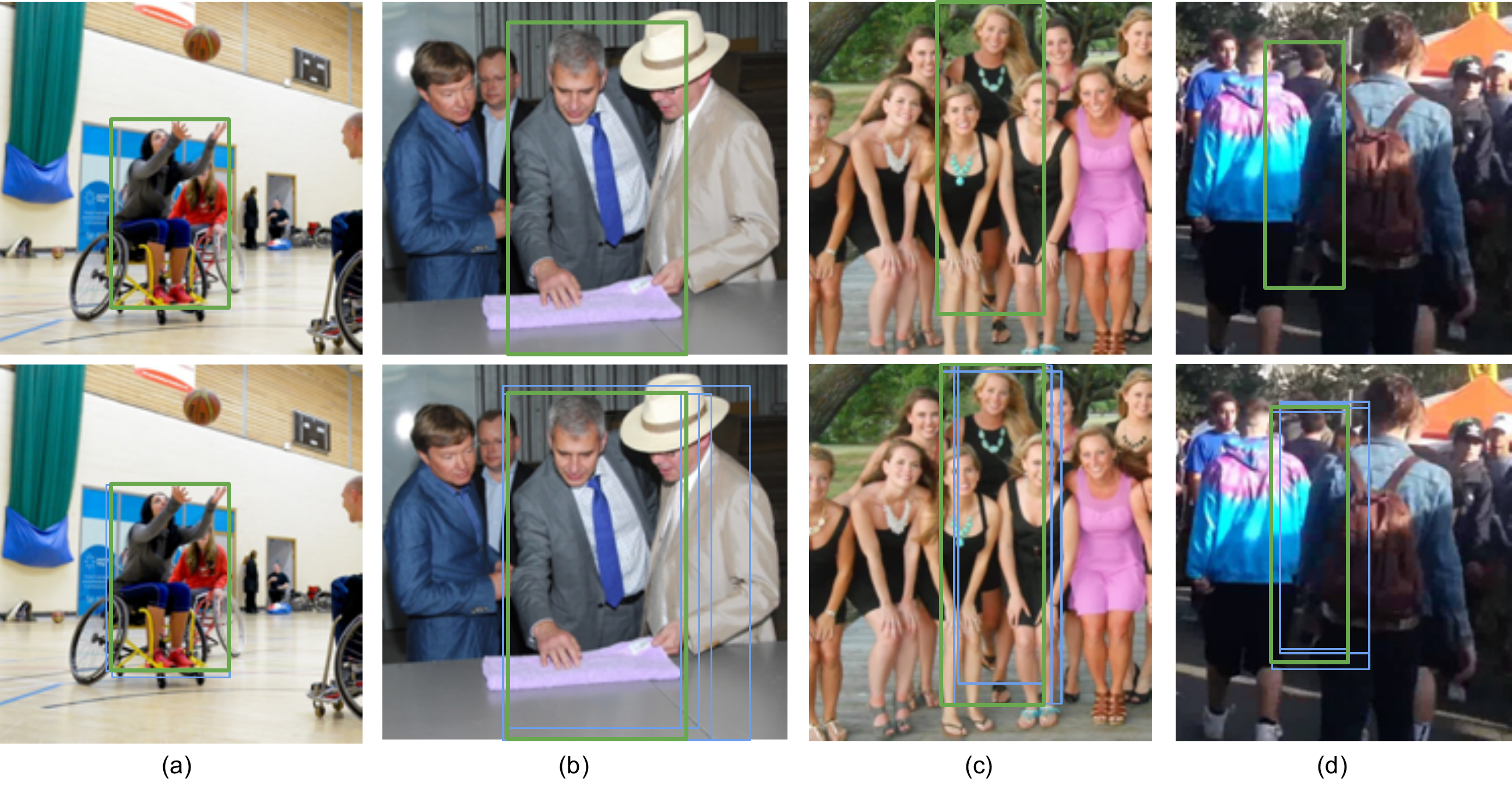}
    \caption{Mixture of Gaussians predictions. 
    First row: baseline Faster R-CNN. 
    Second row: mixture of four Gaussians. 
    {\color{Blue} Blue boxes} are the mixture components. 
    {\color{citecolor} Green boxes} are the final predictions. 
    (a) not occluded (b) left arm is occluded (c) both arms are occluded (d) heavily occluded.
    }
    \label{fig:mix}
\end{center}
\end{figure*}

\begin{figure*}[t]
\begin{center}
\includegraphics[width=\linewidth]{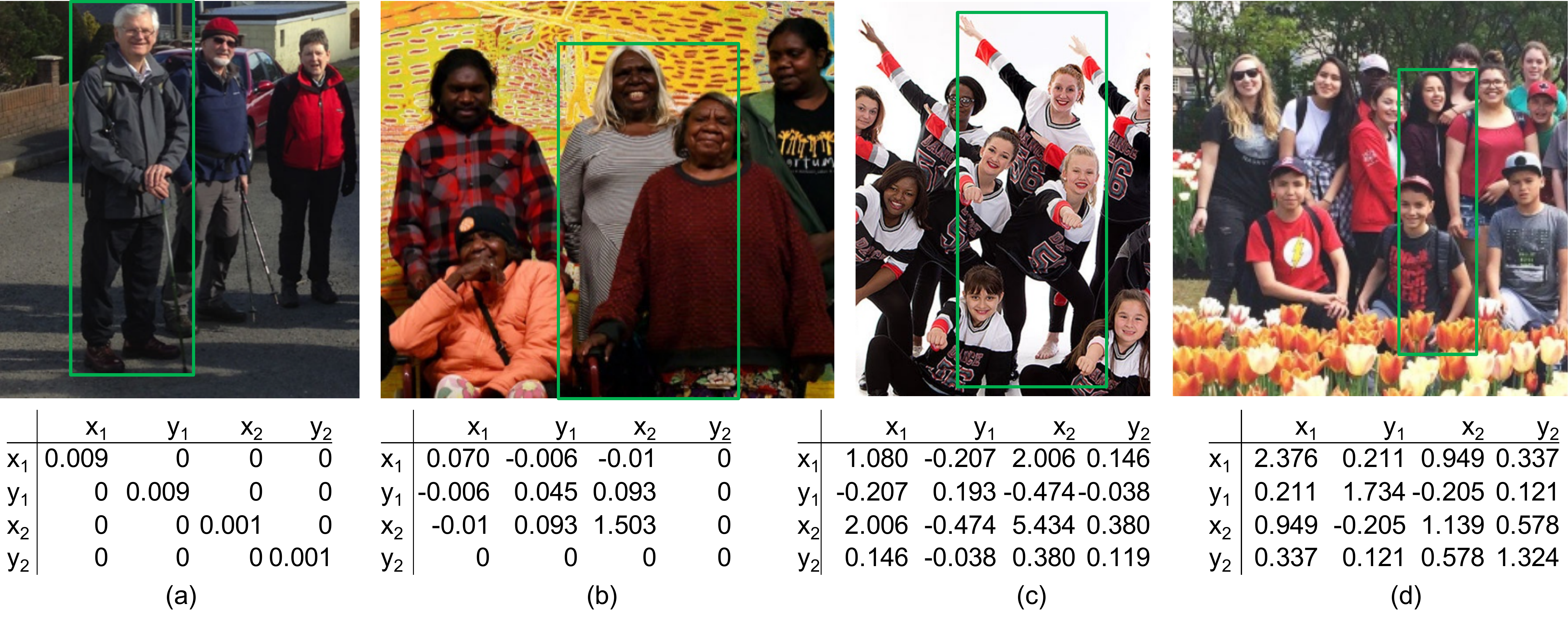}
    \caption{Multivariate Gaussian predictions and corresponding covariances on CrowdHuman. 
    }
    \label{fig:sigma}
    \end{center}
\end{figure*}



\paragraph{Comparison with State-of-the-art} Shown in Table~\ref{tab:deepvoting}, we compare multivariate mixture of eight Gaussians model to DeepVoting~\cite{deepvoting} on VehicleOcclusion. Similar to CrowdHuman, the bounding box annotations are full boxes. The baseline is VGG-16 Faster R-CNN.

Our multivariate mixture of eight Gaussians model outperforms DeepVoting by a large margin at different occlusion levels. Without occlusion, our model also helps to learn a better detector, coinciding the experiments above. We argue that our model considers multiple modes of an object and the correlations between each border of a bounding box, which helps detection under occlusion.

\paragraph{Model Size and Inference Speed} We measure the inference speed of our models using ResNet-50 FPN Mask R-CNN with a TITAN Xp, CUDA 10.1 and cuDNN 7.5.0 on MS-COCO val2017~\cite{cp,amc,phan2020mobinet,liang2017single}. Shown in Table~\ref{tab:time}, Gaussian (Equation~\ref{eq:single}) and multivariate Gaussian (Equation~\ref{eq:multivariate}) neither slow down the inference nor increase the number of parameters, since we can discard the covariance $\bm{\Sigma}$ at inference time (Section~\ref{sec:inference}). 
The complete model, multivariate mixture of eight Gaussians (Equation~\ref{eq:model}), only increases \textbf{2M} parameters and sacrifices \textbf{0.9 FPS} on GPU. Our models outperform the baselines by large margins (Table~\ref{tab:crowdhuman}, \ref{tab:coco} and \ref{tab:deepvoting}), while requires almost no additional computation and memory. 

Note that we measure the inference latency on MS-COCO where there are 80 classes, such that the number of parameters for $\bm{\mu}$ is  $1024\times 80 \times K$ (1024 is the number of output channels of fc7, Figure~\ref{fig:arch}). On CrowdHuman where there is only one class (human), the number of parameters for $\bm{\mu}$ is only $1024 \times K$, which will consume even fewer computation and memory resources. 

\paragraph{probabilistic detection challenge} \cite{hall2020probabilistic} proposed a new metric (PDQ) for measuring probabilistic detection quality on MS-COCO val2017. We follow \cite{hall2020probabilistic} and trained a ResNeXt FPN Faster RCNN with our multivariate Gaussian model (probable, Table~\ref{tab:coco}). Shown in Table~\ref{tab:evaluation}, our approach outperformed \cite{hall2020probabilistic} by a large margin in every metric.  

\subsection{Qualitative Results}
\paragraph{Mixture of Gaussians}
Figure~\ref{fig:mix} shows the visualization of our mixture of Gaussian prediction results on CrowdHuman. When the object is not occluded, our model usually only exhibits a single mode. In Figure~\ref{fig:mix}~(a), the predictions of the mixture components for the athlete are almost the same. When the object is occluded, the occluded bounding box border usually exhibits multiple modes. For example, the left arm of the man can have several reasonable poses in Figure~\ref{fig:mix}~(b).

\paragraph{Multivariate Gaussian}
Figure~\ref{fig:sigma} shows the visualization of our multivariate Gaussian prediction results on CrowdHuman. When the object is not occluded, like in Figure~\ref{fig:sigma}~(a), most terms in the covariance matrix are usually almost zeros. When a border of the object is occluded, like in Figure~\ref{fig:sigma}~(b), the variance term for that border tends to be very high. Sometimes our model learns the covariance between bounding box borders. For example, in Figure~\ref{fig:sigma}~(c), $x_1$ and $x_2$ has a positive correlation, which suggests if the left border moves right, the right border might also move right. When the object is heavily occluded, most of its variance terms are usually very high, shown in Figure~\ref{fig:sigma}~(d).


\section{Conclusion}
We propose a multivariate mixture of Gaussians model for object detection under occlusion. Quantitatively, it demonstrates consistent improvements over the baselines among MS-COCO, PASCAL VOC, CrowdHuman, and VehicleOcclusion. Qualitatively, our model enjoys explainability as the detection results can be diagnosed via the covariance matrices and the mixture components. 


\nocite{choi2019gaussian,zhu2019feature, 8885685, Wang_2020_WACV,he2020epipolar,he2019addressnet}

\bibliographystyle{IEEEtran}
\bibliography{IEEEabrv,IEEEexample}

\end{document}